\documentclass[journal,twoside,web]{ieeecolor}
\pdfoutput=1
\usepackage{generic}
\usepackage{cite}
\usepackage{amsmath,amssymb,amsfonts}
\usepackage{graphicx}
\usepackage{textcomp}
\usepackage{algorithm, algorithmic}
\usepackage{subfigure}
\usepackage{multirow}
\usepackage{utfsym}
\usepackage{booktabs}
\usepackage{threeparttable}
\usepackage{subcaption}
\def\BibTeX{{\rm B\kern-.05em{\sc i\kern-.025em b}\kern-.08em
    T\kern-.1667em\lower.7ex\hbox{E}\kern-.125emX}}
\begin{document}
\title{TCDformer-based Momentum Transfer Model for Long-term Sports Prediction}
 
\author{Hui~Liu, Jiacheng~Gu, Xiyuan~Huang, Junjie~Shi, Tongtong~Feng  and Ning~He
\thanks{This work is supported by the National Natural Science Foundation of China under Grant 62272049, 62236006, 62172045.}
\thanks{H. Liu and X. Huang are with the School of Urban Rail Transit and Logistics, Beijing Union University, Beijing, 100101, China (e-mail: liuhui@buu.edu.cn, xiyuanhuang888@outlook.com). }
\thanks{J. Gu, J. Shi and N. He is with the Smart City College, Beijing Union University, Beijing, 100101, China (e-mail: shijunjie@buu.edu.cn, gujiacheng{\_}04@outlook.com, xxthening@buu.edu.cn). }
\thanks{T. Feng is with the Department of Computer Science and Technology, Tsinghua University, Beijing 100084, China (e-mail: fengtongtong@tsinghua.edu.cn).}}
\maketitle

\begin{abstract}
Accurate sports prediction is a crucial skill for professional coaches, which can assist in developing effective training strategies and scientific competition tactics. Traditional methods often use complex mathematical statistical techniques to boost predictability, but this often is limited by dataset scale and has difficulty handling long-term predictions with variable distributions, notably underperforming when predicting point-set-game multi-level matches. To deal with this challenge, this paper proposes TM$^2$, a TCDformer-based Momentum Transfer Model for long-term sports prediction, which encompasses a momentum encoding module and a prediction module based on momentum transfer. TM$^2$ initially encodes momentum in large-scale unstructured time series using the local linear scaling approximation (LLSA) module. Then it decomposes the reconstructed time series with momentum transfer into trend and seasonal components. The final prediction results are derived from the additive combination of a multilayer perceptron (MLP) for predicting trend components and wavelet attention mechanisms for seasonal components. Comprehensive experimental results show that on the 2023 Wimbledon men’s tournament datasets, TM$^2$ significantly surpasses existing sports prediction models in terms of performance, reducing MSE by 61.64{\%} and MAE by 63.64{\%}.
\end{abstract}


\section{introduction}
\IEEEPARstart{S}{ports} can keep in touch with friends and family, fill spare time, experience the immersive tension, and enjoy the joy of winning, including famous Olympic Games, FIFA World Cup, NBA, the Championships of Wimbledon, and have become inseparable from people's daily life. Analysis from Technavio reports\footnote{https://www.technavio.com/sample-report/sporting-events-market.} shows global sporting events market size is estimated to grow by USD 107.28 billion between 2024 and 2028 with 22.66{\%} annual growth rate, which underscores the immense economic impact of sporting fixtures on societies worldwide. Statistics from Iccopr reports\footnote{https://iccopr.com/wp-content/uploads/2019/03/Sports-Around-the-World-report.pdf} shows that there has been an increasing focus on achieving improved physical and mental well-being. Participation in healthy sports activities and the promotion of different athletic events has expanded dramatically. There are 3.14 billion users who regularly participate in physical exercise around the world in 2025, equating to 50.4{\%} of the total global population aged 60 and below\footnote{https://datareportal.com/social-media-users/.}. According to the KPMG reports\footnote{https://assets.kpmg.com/content/dam/kpmg/cn/pdf/zh/2021/09/olympic-economics-and-sports-industry-outlook.pdf}, 2.16 coaches are needed for every 1,000 sports user. Therefore, professional sports coaching will become a super-emerging and popular profession.

\begin{figure*}[t]
    \centering
    \includegraphics[width=0.7\textwidth]{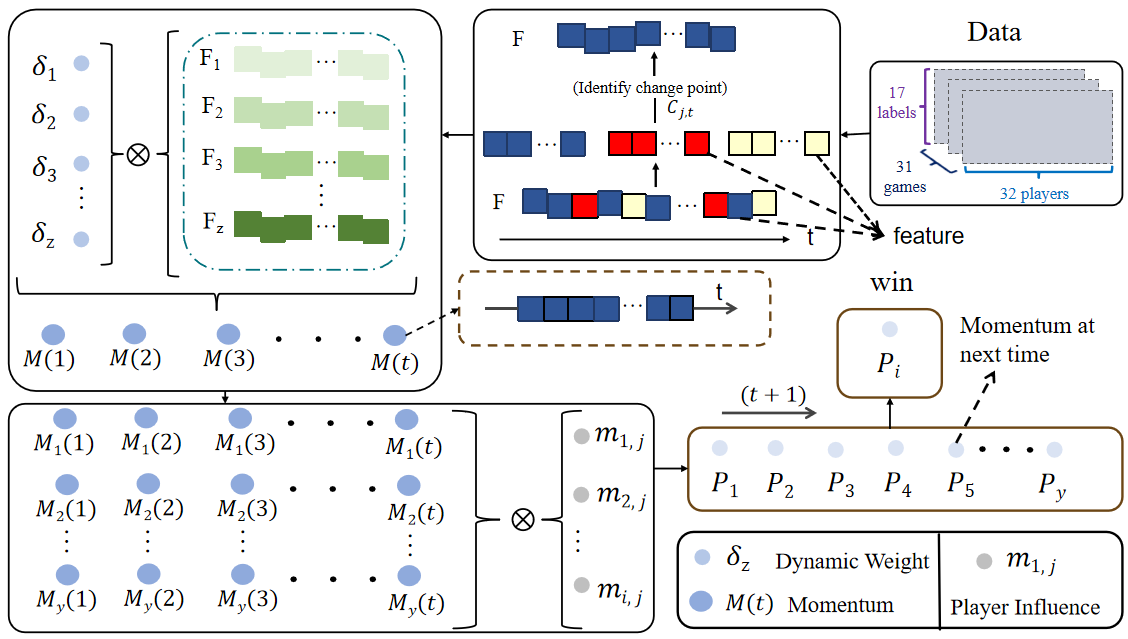}
    \caption{An overview of TCDformer-based momentum transfer model for long-term sports prediction.}
    \label{fig1}
\end{figure*}

Accurate sports prediction\cite{xu2024rethinking} is a crucial skill for professional physical education instructors and coaches because this skill helps them develop effective training strategies and scientific competition tactics, and make wise real-time strategy optimization during games. Specifically, coaches can thoroughly review students’ or athletes’ sports states and mentality changes, assess the strengths and weaknesses of their teams and opponents,  tailor their coaching techniques, and set more realistic and attainable competition objectives accordingly. Additionally, accurate sports prediction can help prevent injuries by adjusting workloads and identifying high-risk situations, contributing to the overall well-being of athletes.

Existing sports prediction models can be categorized into three types: knowledge-based models, data-based models, and knowledge-data-driven models.{\it Knowledge-based models}\cite{pu2024orientation, liao2021softwarized} rely on subjective assessments drawn from the expertise and experience of physical education instructors or coaches. These models make predictions by focusing on key scenarios in a match—such as spatial advantages and critical events—followed by tailored quantitative physical instruction based on the student’s circumstances. They are particularly useful when analyzing every variable in detail is not feasible. However, these models heavily depend on the individual’s expertise and are not easily interpretable, leading to potential biases and limited accuracy. {\it Data-based models}\cite{neumann2023extending, kjamilji2024privacy, mao2023self} use machine learning techniques to provide detailed analyses of specific indicators, player performance, and team dynamics. This approach includes basic prediction techniques like classification and regression modeling, as well as weighted averages, enhanced with statistical analysis to forecast sports outcomes. These models are interpretable and maintain good accuracy for single-game predictions. Nonetheless, with the increasing availability of large-scale data and advanced mathematical processing technologies, data-based models have become more complex, making their practical application in the sports arena challenging. {\it Knowledge-data-driven models}\cite{ullah2021ai, alfarizi2022optimized} represent a hybrid approach that combines the contextual insights of knowledge-based models with the explainable analysis capabilities of data-driven models, such as gray correlation analysis, which merges qualitative and quantitative prediction methods. This integration mitigates the limitations of each individual approach, providing predictions that are grounded in expert knowledge and aligned with the laws of physics and common sense, while still maintaining interpretability. These models are particularly useful in scenarios where a balance between empirical data and expert knowledge is needed. 

However, they are constrained by the fact\cite{zhang2021spovis} that data-driven models often use complex mathematical statistical techniques to boost predictability, but this often is limited by dataset scale and has difficulty handling long-term predictions with variable distributions, notably underperforming when predicting point-set-game multi-level matches. Developing methods for long-term sports predictions using large-scale unstructured time series remains a significant challenge, requiring a trade-off between complexity and interoperability.

Momentum\cite{liang2024foundation} in sports can accurately predict key moments from long-term large-scale unstructured time series based on expert experience, athlete psychology, statistical models, etc. Therefore, the quantitative judgment of momentum in sports can handle this challenge well. Current momentum research\cite{rasul2023lag,garza2023timegpt} mainly focuses on understanding the internal mechanisms and effects. Statistical analysis\cite{wan2024tcdformer} shows that strategic and psychological momentum contributes to the outcome of a match and affects the point, set, and game process. In particular, psychological momentum, rather than strategic momentum, was the main factor in improved player performance after the successful conversion of break points; has a significant prior point or hold success and current match-winning percentage; interruptions in the match had a negative effect on psychological momentum. However, existing methods lack accurate momentum encoding and explainable prediction models based on momentum.

To deal with this challenge, this paper proposes TM$^2$,(see in Fig. \ref{fig1}) a TCDformer-based Momentum Transfer Model based on the theoretical foundations of wavelet transformation, attention mechanisms, and Transformer networks. TM$^2$ encompasses two key modules: a momentum encoding module and a prediction module based on momentum transfer. TM$^2$ initially encodes momentum in large-scale unstructured time series using the local linear scaling approximation (LLSA) module. Then it decomposes the reconstructed time series with momentum transfer into trend and seasonal components. The final prediction results are derived from the additive combination of a multilayer perceptron (MLP) for predicting trend components and wavelet attention mechanisms for seasonal components. Comprehensive experimental results show that on the 2023 Wimbledon men’s tournament datasets, TM$^2$ significantly surpasses existing sports prediction models in terms of performance, reducing MSE by 61.64{\%} and MAE by 63.64{\%}. This approach offers an effective framework for managing large-scale unstructured long-term time series, achieving a balance between performance and interpretability.

\begin{table*}[t]%
\centering
\fontsize{9pt}{12pt}\selectfont
\caption{Existing advanced sports event prediction models} \label{tab1}
\begin{threeparttable}
\begin{tabular*}{\textwidth}{c p{14cm}c}
\toprule
\multicolumn{1}{c}{\textbf{Author}}   & \multicolumn{1}{c}{\textbf{Modeling Technique(s) and Features}}  & \multicolumn{1}{c}{\textbf{Years}} \\
\midrule
CF-LSTM \cite{xu2020cf}   &Integrate the feature information of the pedestrians from the first two time steps into a separate input to the LSTM and focus on the internal features of the dynamic interactions& 2020\\
DNRI \cite{graber2020dynamic}   &Formulate explicit recovery of system interactions as NRI of latent variables& 2020\\
DMA-Nets \cite{ji2021dynamic}   &End-to-end RNN-based model with a hierarchical dynamic attention layer is introduced that uses two temporal attention mechanisms to enhance the model's ability to represent complex conditional dependencies in real-world datasets, while the temporal prediction layer ensures that predicted citations are monotonically increasing along the temporal dimension& 2021\\
Seq2Event \cite{simpson2022seq2event} &A Combined Model of Recurrent Neural Networks (RNN), Long Short-Term Memory (LSTM), Gated Recurrent Units (GRUs) and Transformers & 2022 \\
ShuttleNet \cite{wang2022shuttlenet} & Neural Network with have two encoder-decoder extractors and a fusion network)  &2022 \\
\bottomrule 
\end{tabular*} 
\end{threeparttable} 
\end{table*}

The rest of this paper is organized as follows. Section \uppercase\expandafter{\romannumeral2} briefly reviews the related works Section \uppercase\expandafter{\romannumeral3} presents the overview and design details of TM$^2$.
Section \uppercase\expandafter{\romannumeral4} presents the implementation details of TM$^2$. Section \uppercase\expandafter{\romannumeral5} presents the evaluation results compared to existing models. Conclusion and future works are discussed in Section \uppercase\expandafter{\romannumeral6}.

\section{Related Works} \label{sec2}

In this section, we examine the existing research on sports event prediction models and time series prediction models, two key areas relevant to our study.

\subsection{Sports Event Prediction Models}

With the rise of machine learning and deep learning, sports prediction models have evolved significantly, employing various data-driven and knowledge-based methods\cite{liao2021softwarized}. Zhiqiang Pu et al.\cite{pu2024orientation} categorized these models into three types: knowledge-based models, data-driven models, and integrated knowledge-data-driven models, each with specific applications and limitations.TABLE \ref{tab1} shows the best models in recent years.

\subsubsection{Knowledge-Based Models} Sports events are inherently complex, making it challenging to analyze every aspect comprehensively. Researchers often focus on key scenarios, such as spatial advantages in a match or crucial game events. For instance, in tennis, serve analysis is a prominent research area since serving is the most frequent event in the game. These models rely on expert knowledge\cite{zhang2023new} to evaluate specific in-game factors and focus on scenario-based analysis, offering valuable insights but limited scalability due to their reliance on expert understanding.

\subsubsection{Data-Driven Models} The advancement of machine learning techniques\cite{neumann2023extending, kjamilji2024privacy, mao2023self} and the availability of large datasets have expanded the scope of sports event analysis. Common data types include event data and tracking data. Models built on event data focus on metrics such as expected wins, serve analysis, and player performance\cite{pang2022hybrid} while tracking data is used for more comprehensive evaluations, including tactical assessments and team dynamics. These data-driven approaches provide a deeper understanding\cite{zhang2021spovis} of player and team performance but often lack interpretability due to their complexity.

\subsubsection{Knowledge-Data-Driven Models} To address the interpretability challenges of data-driven models, integrated knowledge-data-driven approaches\cite{ullah2021ai} have emerged. These models combine expert knowledge with machine learning techniques, leveraging the strengths of both. Over time, sports event prediction has shifted from simple statistical models to advanced machine learning models that integrate domain expertise\cite{alfarizi2022optimized}. This trend reflects the increasing demand for models that are both interpretable and data-driven, which are more practical for coaches, analysts, and players seeking actionable insights.

\subsection{Time Series Prediction Models}

Time series data presents unique challenges due to its dynamic temporal structure and varying patterns across domains. Recent work by Yuxuan Liang et al.\cite{liang2024foundation} categorizes time series models into three main types: standard time series models, spatial time series models, and other temporal data models, each with distinct applications and strengths.

\subsubsection{Standard Time Series Models} Standard time series models are designed to capture general patterns from large datasets, typically aimed at forecasting or classification tasks. These models are often pre-trained on vast amounts of time series data across various domains. For example, Lag-Llama \cite{rasul2023lag} uses a decoder-only transformer architecture, while TimeGPT-1 \cite{garza2023timegpt} adopts an encoder-decoder structure with transformer layers. These models focus on task-specific improvements and are resource-intensive to train from scratch. Other approaches, such as LLM4TS \cite{chang2023llm4ts} and TEMPO \cite{cao2023tempo}, successfully fine-tune large language models for time series forecasting, demonstrating the adaptability of pre-trained models to non-linguistic data.TTM \cite{ekambaram2024ttms}, which aims to create domain-agnostic prediction models, and Moirai \cite{woo2024unified}, which introduces a universal predictive transformer based on a mask encoder that is pre-trained on a large dataset (LOTSA) containing observations from different domains.

\begin{figure*}[t]
\centering
\includegraphics[width=0.85\textwidth]{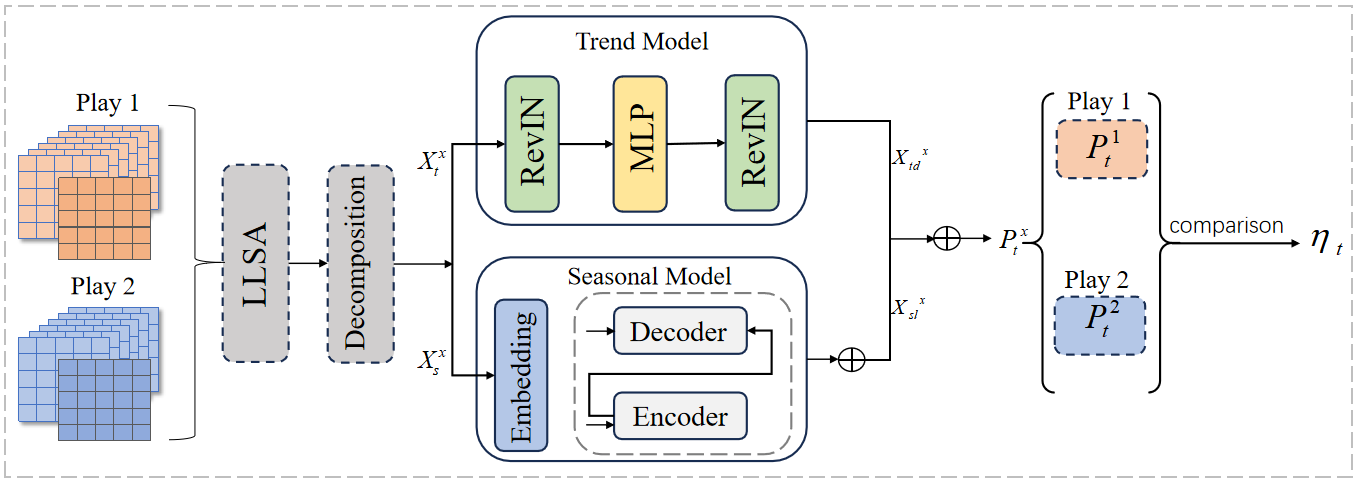}
\caption{Model result output process based on TCDformer.}
\label{fig2}
\end{figure*}

\subsubsection{Spatial Time Series Models} Many real-world systems generate spatial-temporal data\cite{feng2024u2udata}, necessitating models that account for spatial dependencies alongside temporal dynamics. These spatial time series models often utilize graph structures or raster data to represent the spatial component. In transportation, for example, models like TFM \cite{wang2023building} analyze urban traffic patterns using graphs, while DiffSTG \cite{wen2023diffstg} employs diffusion models for probabilistic forecasting. In the field of climate science, models such as FourCastNet \cite{pathak2022fourcastnet} and Pangu-Weather \cite{bi2023accurate} use spatial-temporal grids to make accurate global weather predictions, outperforming traditional numerical methods.

\subsubsection{Other Temporal Data Models} Beyond traditional time series and spatial-temporal data, other domains such as trajectory data and clinical records also involve temporal dynamics. In human mobility prediction, models like AuxMobLCast \cite{xue2022leveraging} fine-tune large language models to predict movement patterns, while DiffTraj \cite{zhu2023difftraj} reconstructs geographic trajectories using diffusion processes. These models illustrate the flexibility of time series methodologies across various industries and their increasing importance in forecasting and decision-making processes.

\section{methodology}\label{sec3}

\subsection{Momentum Encoding Module}

Given the similarity between change points and trends in non-stationary time series, ignoring the impact of potential change points (specifically, apparent fluctuations driven by external events rather than simple noise) may lead to misleading conclusions. To reduce this risk, we apply the local linear scaling approximation (LLSA) module proposed by Jiashan Wan et al. \cite{wan2024tcdformer} to encode momentum in large-scale unstructured time series(see in Fig. \ref{fig2}). The specific steps are as follows:

\subsubsection{Step 1: Extraction of Wavelet Coefficients}
First, let $X$ represent the time series data with dimensions $T \times D$, where $T$ denotes the length of the series and $D$ represents the number of variables. We define $X$ as $X = (x_{1}, \ldots, x_{l}, \ldots, x_{T})^{T} \in \mathbb{R}^{T \times D}$, where each $x_{t}$ (for $1 \leq t \leq T$) is a $D$-dimensional vector representing the values of all $D$ variables at time $t$, written as $x_{t} = (x_{1}^{d}, \ldots, x_{l}^{d}, \ldots, x_{T}^{d})^{T} \in \mathbb{R}^{D}$. Similarly, for each variable $d$ (where $1 \leq d \leq D$), the series $x^{d} = (x_{1}^{d}, \ldots, x_{l}^{d}, \ldots, x_{T}^{d})^{T} \in \mathbb{R}^{T}$ describes its temporal evolution across all $T$ time points.

Next, we apply the MODWT to the time series $X$ to obtain the wavelet coefficients. These coefficients represent the differences between moving averages at different scales $\widetilde{s_{j}}$:
\begin{equation}
    \left\{
    \begin{aligned}
        l_{j} &= \arg \max_{t} (|W_{j,t}|) \\
        l_{min}^{j} &= \min \{ t \mid t \in \sup(W_{j}) \} \\
        l_{max}^{j} &= \max \{ t \mid t \in \sup(W_{j}) \}
    \end{aligned} \tag{1}
    \label{eq:1}
    \right.  
\end{equation}
where $W_{j}$ denotes the set of wavelet coefficients obtained through MODWT, $\arg \max_{t}$ identifies the value of $t$ that maximizes the given condition, and $l_{j}$ corresponds to the $t$ value at which $|W_{j,t}|$ reaches its maximum. The notation $\sup(W_{j})$ indicates the set of positions in $W_{j}$ that contain non-zero elements, while $l_{min}^{j}$ and $l_{max}^{j}$ represent the positions of the minimum and maximum non-zero elements of $t$ within this set, respectively.

\subsubsection{Step 2: Identification and Characterization of Change Points}
Change points are identified by observing the sign changes in the wavelet coefficient vectors on either side of the detected change points. These sign changes are key for determining the extent of the change. The changes in sign on either side of the change point in the wavelet coefficient vectors are defined as follows:
\begin{equation}
    \left\{
    \begin{aligned}
        n_{\alpha,j} &= \sum_{t=l_{max,j}}^{l_{j}-1} \frac{|W_{j,t+1} - W_{j,t}|}{2} \\
        n_{\beta,j} &= \sum_{t=l_{j+1}}^{l_{max,j}} \frac{|W_{j,t+1} - W_{j,t}|}{2}
    \end{aligned} \tag{2}
    \label{eq:2}
    \right.
\end{equation}
starting with the wavelet coefficient that has the highest absolute value, which signifies the precise change location, the coefficients at the change points are typically represented by the sign changes in the coefficient vectors on both sides. These are denoted as $n_{\alpha,j}$ and $n_{\beta,j}$, but due to their invariance across all wavelet types and transformation orders, they can also be denoted simply as $n_{\alpha}$ and $n_{\beta}$. To reconstruct the entire jump segment, it is necessary to determine the boundaries of the jump. We set $n_{\alpha}$ as the left boundary, $\alpha$, and $n_{\beta}$ as the right boundary, $\beta$, defining $\Omega = [\alpha, \beta]$ as the complete jump range. The specific calculations for $\alpha$ and $\beta$ are given by:
\begin{equation}
    \alpha = \max \{ l \in [1, l-1] \mid \sum_{t=1}^{l-1} \frac{|W_{j,t+1} - W_{j,t}|}{2} \geq n_{\alpha} \}
    \tag{3}
    \label{eq:3}
\end{equation}
\begin{equation}
    \beta = \min \{ l \in [l+1, T] \mid \sum_{t=l+1}^{L} \frac{|W_{j,t-1} - W_{j,t}|}{2} \geq n_{\beta} \}
    \tag{4}
    \label{eq:4}
\end{equation}

\subsubsection{Step 3: Detection of the $k$th Jump at Scale $J$}
To detect subsequent jumps, we apply a detection rule similar to that of the first jump, with the added step of excluding already-detected jumps and their surrounding regions. The position of the $k$th jump is defined as:
\begin{equation}
    l_{k} = \arg \underset{t}{\max} (|W_{J,t}| \mid t \notin \underset{1 \leq i \leq k}{\cup} \Omega_{i})
    \tag{5}
    \label{eq:5}
\end{equation}
in this context, $l_{k}$ represents the position of the $k$th jump, $\Omega_{i}$ denotes the $i$th jump region, and $\cup_{1 \leq i \leq k} \Omega_{i}$ represents the union of all previously identified jump regions. Here, $\arg \max_{t}$ is used to identify the value of $t$ that maximizes $|W_{J,t}|$, assuming that $t$ does not fall within any of the previously detected jump regions.

\begin{center}
\begin{table*}[!h]%
\fontsize{8pt}{12pt}\selectfont
\caption{Pressure values between different players}
\label{tab2}
\begin{tabular*}{\textwidth}{cccccccc}
\toprule
\textbf{Player} & \textbf{Carlos Alcaraz} & \textbf{Alexander Zverev}& \textbf{Frances Tiafoe} & \textbf{· \ · \ ·} & \textbf{David Goffin } & \textbf{Maximilian Marterer} & \textbf{Novak Djokovic}\\
\midrule
Alexander Zverev    & 1                & 4.78           & 3.79           & · \ · \ · & 0.71         & 3.32                & 1.27           \\
Carlos Alcaraz      & 0.21             & 1              & 4.85           & · \ · \ · & 1.72         & 4.23                & 2.11           \\
Frances Tiafoe      & 0.26             & 0.21           & 1              & · \ · \ · & 3.03         & 3.97                & 2.72           \\
· \ · \ ·           & · \ · \ ·        & · \ · \ ·      & · \ · \ ·      & · \ · \ · & · \ · \ ·    & · \ · \ ·            & · \ · \ ·            \\
David Goffin        & 1.41             & 0.58           & 0.33           & · \ · \ · & 1            & 1.25                & 3.82           \\
Maximilian Marterer & 0.31             & 0.24           & 0.25           & · \ · \ · & 0.8          & 1                   & 3.63           \\
Novak Djokovic      & 0.79             & 0.47           & 0.37           & · \ · \ · & 0.26         & 0.28                & 1              \\ 
\bottomrule
\end{tabular*}
\end{table*}
\end{center}

\subsubsection{Step 4: Detection at Reduced Scale $j < J$}
After identifying the jump regions at scale $J$ as $\Omega_{k}$, for $1 \leq k \leq K$, the scale is reduced to $J - \Lambda \leq j \leq J$, and $l_{j,k}$ is determined, where $0 \leq \Lambda \leq J$ dictates which scale’s details are reconstructed:
\begin{equation}
    l_{j,k} = \arg \underset{t}{\max} (|W_{J,t}| \mid t \in \Omega_{j+1,k})
    \tag{6}
    \label{eq:6}
\end{equation}
this ensures consistent jump detection across different scales, with the range covering these jumps at scale $j$ denoted as $\Omega_{j,k} = [\alpha_{j,k}, \beta_{j,k}]$.

\subsubsection{Step 5: Signal Reconstruction}
Once the detection of jump positions across all regions $\Omega_{j,k}$ is complete, we perform the inverse MODWT for $1 \leq j \leq J$ to obtain the wavelet coefficients $\widetilde{W}_{j,t}$ that contain the jump information. These are then used to compute the modified $\widetilde{D}_{j,t}$, ultimately leading to the reconstruction of the signal $\overline{X}$.

\begin{figure}[t] 
	\centering
	\includegraphics[width=0.4\textwidth]{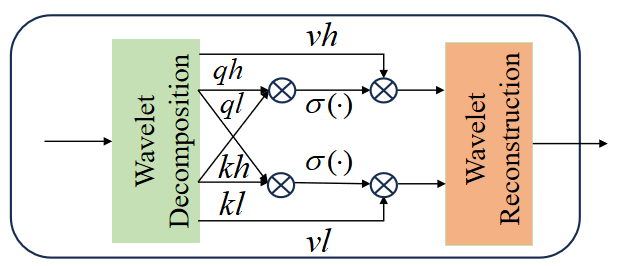}
	\caption{Wavelet attention structure diagram}
	\label{fig3}
\end{figure}
	
\subsection{Prediction Module Based On Momentum Transfer}

To obtain the weights, we first use the analytic hierarchy process (AHP) to evaluate the importance of various indicators and derive the relative importance of the $z$th indicator relative to other indicators, $g_z$, and the degree of influence of the opponent, $\overline{g_z}$. The following formula is used to calculate the weighted value of each indicator over time:

\begin{equation}
\delta_z(t) = d \cdot g_z \cdot \log\left((r_z^t)^{k}\right) + \frac{1}{t+1} \tag{7}
\label{eq:7}
\end{equation}
\begin{equation}
\overline{\delta_z(t)} = d \cdot \overline{g_z} \cdot \log\left((r_z^t)^{k}\right) + \frac{1}{t+1} \tag{8}
\label{eq:8}
\end{equation}
where $\delta_z(t)$ represents the weighted value of the $z$th metric at time $t$, $\overline{\delta_z(t)}$ represents the impact of the $z$th metric on the player when the opponent is paired at time $t$. $r_z^t$ represents the original value of the metric, $d$ is the limiting factor, reflecting the growth constraint, $k$ is the gap between the same type of data in the metric, and different features may need to square the value or use the original value according to their specific properties.

The weights for each point calculated using the above method are used to calculate the player's momentum at each time point. By multiplying these weights by each eigenvalue at each time point (denoted as $w_{1}$ for player 1 and $w_{2}$ for player 2) and summing the products, the player's momentum $M_y(t)$ at a specific moment is determined as follows:
	
\begin{equation}
    M_y(t) = m_{i,j} \cdot \sum\limits_{z=0}^{n}\left(\delta_z(t) \cdot \overline{X}_{i,z}(t) - \overline{\delta_z(t)} \cdot \overline{X}_{j,z}(t)\right) \tag{9}
    \label{eq:9}
\end{equation}
where $M_y(t)$ represents the momentum of player $y$ at time $t$, $m_{i,j}$ is the pressure value of each player facing other players (see in TABLE \ref{tab2}), $\overline{X}{i,z}$ and $\overline{X}{j,z}$ represent the data of player $i$ and player $j$, respectively, reconstructed by Local Linear Seasonal Adjustment (LLSA) based on Maximal Overlap Discrete Wavelet Transform (MODWT). $n$ is the total number of features, while $\delta_z(t)$ and $\overline{\delta_z(t)}$ are the influence weights of the $z$th feature on the player and opponent, respectively.

\begin{table*}[t]%
\centering
\fontsize{9pt}{12pt}\selectfont
\caption{Retained model variables}
\label{tab3}
\begin{threeparttable}
\begin{tabular*}{0.86\textwidth}{cc}
\toprule
\multicolumn{1}{c}{\textbf{Targets}}  &
\multicolumn{1}{c} {\textbf{Explanation}} \\
\midrule
elapsed\_time & Time elapsed since the start of the first point to the start of the current point (H: MM: SS)   \\
p\_sets & Sets won by player \\
p\_games & Games won by a player in current set    \\
server & Server of the point    \\
point\_victor & Winner of the point   \\
p\_ace & Player hit an untouchable winning serve    \\
p\_double\_fault & Player missed both serves and lost the point    \\
p\_break\_pt\_missed & One player misses a chance to win the match while the other is serving     \\
p\_break\_pt\_won & One player wins while the other is serving.     \\
p\_distance\_run & Player's distance ran during a point (meters)   \\
psychological\_factor & The psychological impact of a player's gain or loss during a match \\
\bottomrule
\end{tabular*}
\begin{tablenotes}
\footnotesize
\item[$^{\rm a}$] According to the indicators in the table, player 1 and player 2 recorded data separately.
\end{tablenotes}
\end{threeparttable}
\end{table*}

Next, we decompose the reconstructed time series with momentum transfer into trend and seasonal components. The trend component $x_t$ is calculated using multiple averaging filters of different sizes and integrated into the final trend component through adaptive weighting, while the seasonal component $x_s$ is obtained by subtracting the trend component from the original time series:

\begin{equation}
\left\{
\begin{aligned}
x_s &= M_{y} - x_t\\
x_t &= \sigma(w(x) \ast f_2(x))  
\end{aligned}\tag{10}
\label{eq:10}
\right.
\end{equation}
here, $\overline{X}$ represents the data reconstructed via Local Linear Seasonal Adjustment (LLSA) based on Maximal Overlap Discrete Wavelet Transform (MODWT), and $x_s$ and $x_t$ denote the seasonal and trend components, respectively. $\sigma(\cdot)$, $w(x)$, and $f_2(x)$ represent the softmax function, adaptive weights of average filters, and averaging filter, respectively.
	
For trend prediction, a three-layer MLP is utilized, and to address non-stationarity, RevIN normalization is applied before and after the MLP layers:
	
\begin{equation}
\overline{x_t} = RevIN(MLP(RevIN(x_t))) \tag{11}
\label{eq:11}
\end{equation}
	
For the seasonal component, wavelet-based attention mechanisms are applied, where attention calculations are performed on the decomposed queries, keys, and values at each scale. The process is detailed in the following equation:
	
\begin{equation}
Y(q,k,v) = \overline{W}\left(softmax\left(W(q)\overline{W(k)}^{T}\right)W(v)\right) = qk^{T}v \tag{12}
\label{eq:12}
\end{equation}	
the final momentum forecast is obtained by summing the output of the next point of trend and seasonal components:

\begin{equation}
P(t+1) =(\overline{x_{t}}(t+1)+x_{s}(t+1)) \tag{13}
\label{eq:13}
\end{equation}	
where $P(t_{n+1})$ represents the total momentum of each player at time $t+1$. The continuous nature of match updates ensures that total momentum is refreshed with each time increment and scoring event, maintaining accuracy in the outcome prediction.

\subsection{Result Determination Layer}
The final match outcome relies on the comparative analysis of total momentum values, which encapsulate the players' capabilities during the match. This approach inherently reflects the players' on-field prowess and status at specific moments. The resultant total momentum also mirrors a player's confidence level, a critical determinant in matches between players of comparable skill.

Given the continuous update of match data, the total momentum of each player is refreshed with each time increment and scoring event. The final momentum comparison between players is derived from the following equation:

\begin{equation}
	\eta_{t_{n+1}}=\left\{
	\begin{aligned}
		i  &&  {P_{i}(t_{n+1}) > P_{j}(t_{n+1})} \\
		i \ \text{or} \ j  &&  {P_{i}(t_{n+1}) = P_{j}(t_{n+1})} \\
		j  &&  {P_{i}(t_{n+1}) < P_{j}(t_{n+1})}
	\end{aligned}
	\right.
	\tag{14}
	\label{enq:14}
\end{equation}
in scenarios where players exhibit identical momentum, historical rankings serve as the tiebreaker. Although these rankings are not directly incorporated into the model's calculations, leveraging them in such instances offers a pragmatic and often accurate resolution method. While acknowledging that lower-ranked players can occasionally outperform higher-ranked counterparts, the predicted momentum—derived from in-match data—provides a robust and credible basis for determining outcomes.

To conclude, the momentum-based model offers a systematic and dynamic method for assessing match outcomes, relying on real-time performance metrics.This approach ensures that the final match outcome is reflective of the players' real-time capabilities and their dynamic performance during the match.

\section{evaluation}\label{sec4}

\subsection{Methodology}

\begin{table*}[!h]%
\fontsize{9pt}{12pt}\selectfont
\centering
\caption{Model comparison results of TM$^{2}$}
\label{tab4}
\begin{tabular*}{0.8\textwidth}{p{2.5cm}p{1.5cm}p{1.5cm}p{1.5cm}p{1.5cm}p{1.5cm}p{1.5cm}p{1.5cm}}
\toprule
\textbf{Metrics} & \textbf{ELO} & \textbf{DT}& \textbf{LR}&\textbf{SVM} &\textbf{RF} & \textbf{ TM$^{2}$ } \\
\midrule
MAE    &  0.4859  & 0.2644  & 0.2126 &0.2195 &0.2249  &0.0389 \\
MSE    &  0.4859  & 0.2644  & 0.2126 &0.2195 &0.2249  &0.0671 \\
Accuracy &0.5141  &0.7395   &0.7874  &0.7805 &0.7751  &0.9237 \\
Precision &0.2643 &0.7397	&0.7824	 &0.7831 &0.7752  &0.9231 \\
F1-score & 0.3491 &0.7396   &0.7871	 &0.7804 &0.7751  &0.9206 \\
\bottomrule
\end{tabular*}
\end{table*}

\emph{Datasets.} The experiments utilize a public dataset, ETTh1, which contains detailed information from the 2023 Wimbledon tennis tournament. This dataset was obtained through the 2023 International Mathematical Modeling Competition and cross-referenced with data from the official Wimbledon Tennis Championships website. During the data collection process, the metrics were recorded separately for each player at the key moments of each match, rather than being combined. This resulted in a dataset containing 7,285 rows and 49 columns, totaling 356,965 data points. Since many of these data points were not highly relevant to the momentum analysis, dimensionality reduction techniques were applied, reducing the dataset to 18 key features. As shown in TABLE \ref{tab3}, these retained features play a critical role in determining match outcomes.

\emph{Evaluation Metrics.} Given that the primary model utilized is a time series model, Mean Squared Error (MSE) and Mean Absolute Error (MAE) are selected as the primary evaluation metrics. Furthermore, since this is a predictive problem, Accuracy, Precision, and F1-score are also included to provide a comprehensive assessment of prediction accuracy. The F1-score, in particular, is crucial for verifying the balance between precision and recall in the model’s performance.

\emph{Baselines.} We selected several commonly used models in tennis match prediction for comparison, including the ELO rating system\cite{neumann2023extending}, Decision Tree (DT)\cite{kjamilji2024privacy}, and Logistic Regression (LR)\cite{mao2023self}. In addition, two widely-used machine learning algorithms—Support Vector Machine (SVM)\cite{pang2022hybrid} and Random Forest (RF)\cite{alfarizi2022optimized}—were included for comparison. For training, the entire dataset was used to optimize the parameters of each model. To ensure reliable results, 80\% of the data was randomly selected as the training set, while the remaining 20\% was used for testing. Each experiment was repeated 100 times to account for randomness, and the average of the performance metrics was calculated to produce the final results.

\subsection{Performance}
\begin{figure}[t]
	\centering
	\includegraphics[width=0.45\textwidth]{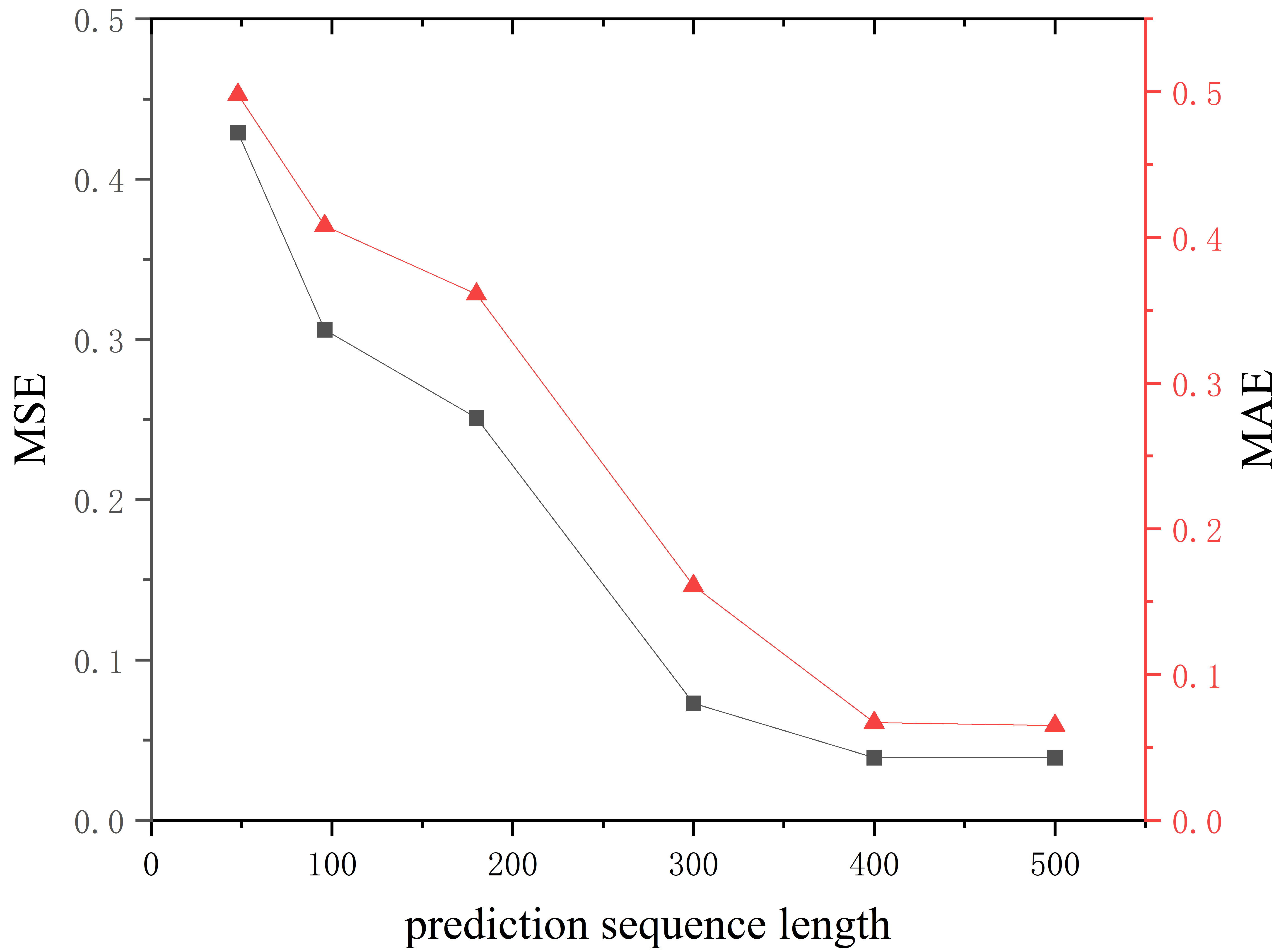}
	\caption{Iterative adjustment curve of TM$^{2}$ prediction sequence length.}
	\label{fig4}
\end{figure}

\emph{Parameter Optimization.} This section discusses the process of parameter optimization for the proposed model, TM$^2$. Since TM$^2$ is a deep learning model, proper parameter tuning is crucial for achieving optimal performance. We used MSE and MAE, common metrics in deep learning, to guide the optimization process. By iteratively adjusting the prediction sequence length, we identified the optimal configuration. To mitigate the effects of outliers, the model was trained 10 times for each sequence length, and the average result was taken. After extensive experimentation, a prediction sequence length of 400 was selected as the optimal setting. The results are presented in Fig. \ref{fig4}.

\begin{figure}[t] 
	\centering
	\includegraphics[width=0.45\textwidth]{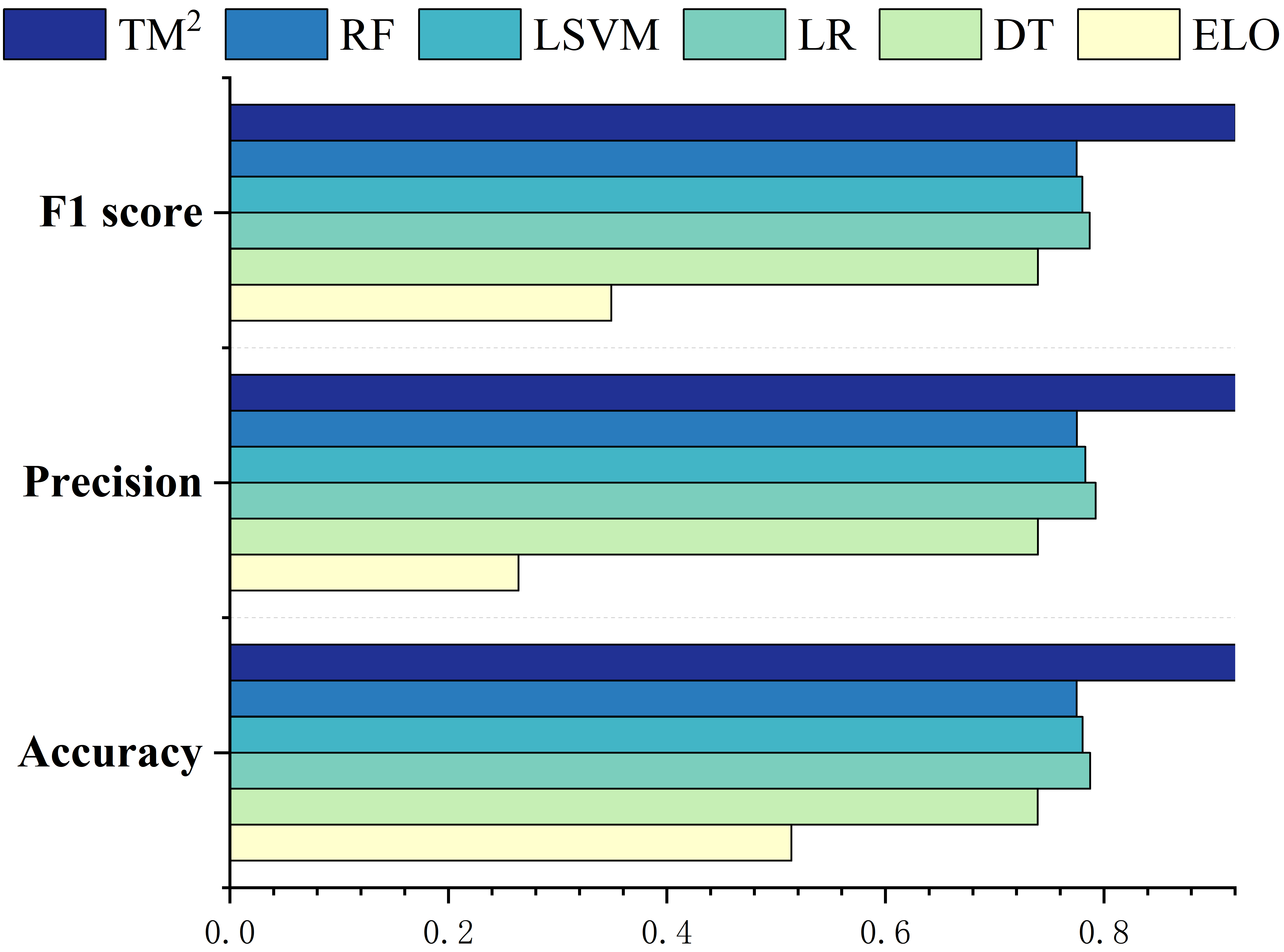}
	\caption{Comparison of TM$^{2}$ with existing basic tennis models in terms of F1-score, precision and accuracy metrics, in the task of predicting tennis match results, with data from the 2023 Wimbledon tournament.}
	\label{fig5}
\end{figure}

\begin{figure}[t] 
	\centering
	\includegraphics[width=0.45\textwidth]{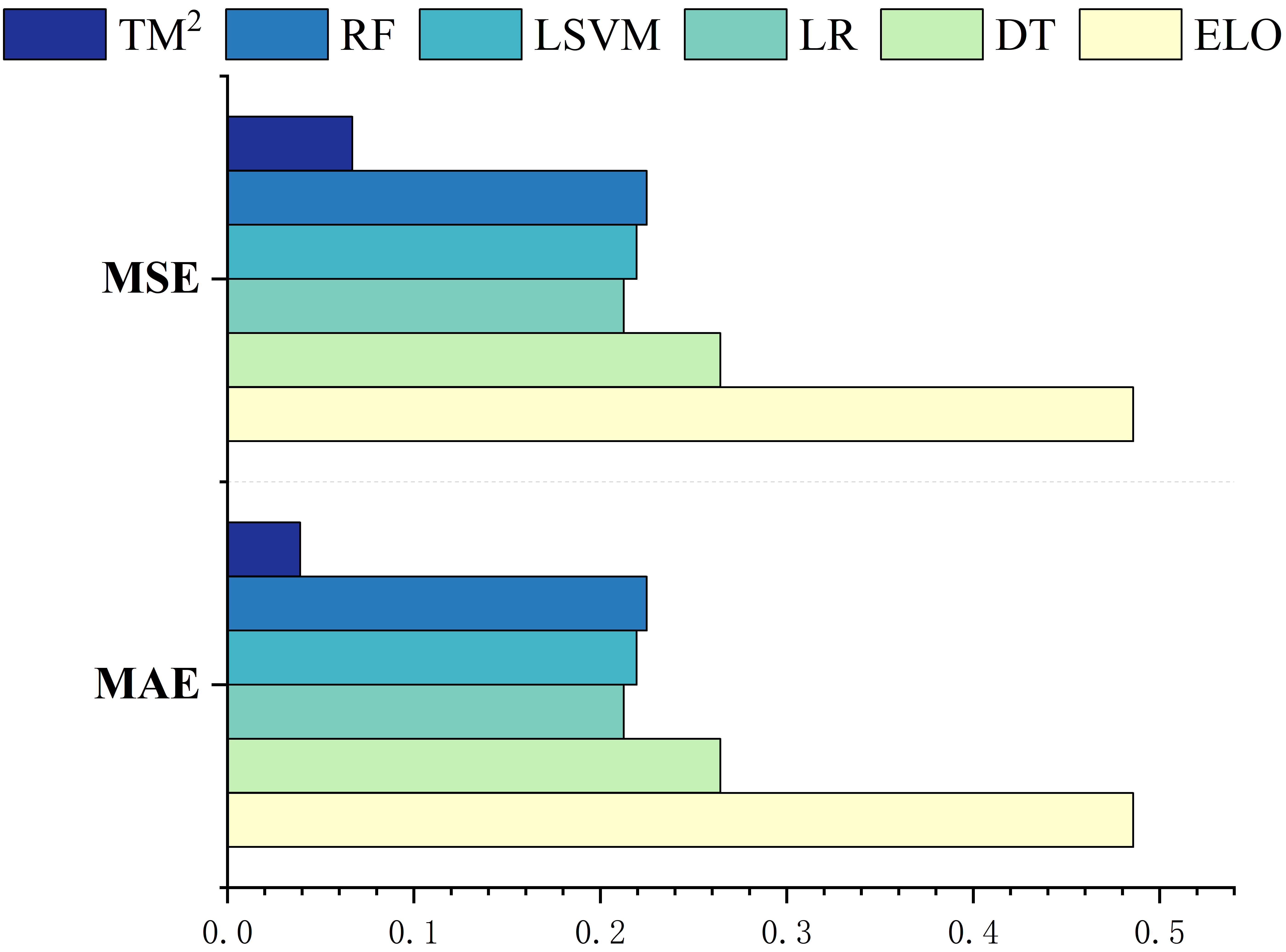}
	\caption{Comparison of TM$^{2}$ with existing basic tennis models in terms of MSE and MAE, in the task of predicting tennis match results, with data from the 2023 Wimbledon tournament.}
	\label{fig6}
\end{figure}

From TABLE \ref{tab4}, Fig. \ref{fig5} and Fig. \ref{fig6}, it is evident that TM$^2$ outperforms the baseline models in terms of Accuracy, Precision, and F1-score. Traditional models like DT, LR, and SVM rely on specific weighting strategies and perform better when data points are independent of each other. However, in scenarios where consecutive data points influence each other, such as in time series data with numerous change points, these models tend to perform poorly, leading to inductive bias in predictions.

TM$^2$ also surpasses the baseline models in terms of MSE and MAE. The MSE and MAE values of TM$^2$ are significantly lower than those of the other models. This is likely because MSE and MAE are more suited for evaluating continuous predictions, which are a key feature of deep learning models. By contrast, Accuracy, Precision, and F1-score, which rely on True Negatives (TN), False Negatives (FN), False Positives (FP), and True Positives (TP), are typically used in classification tasks. Nonetheless, TM$^2$ shows a clear advantage in both types of metrics, highlighting its robustness.

To understand the comparison results of TM$^2$ with existing advanced models, we collected the DMA-Nets model released by Taoran Ji in 2021 and the Seq2Event model published by Ian Simpson in 2022. Both models have excellent prediction results in their respective fields and can represent the development direction of sports event result prediction to a certain extent. Therefore, we compare the results of these two models with our model in MSE and MAE in TABLE \ref{tab5}. It can be found that TM$^2$ is better than DMA-Nets and Seq2Event. This shows that our model has excellent development capabilities and there is room for improvement.

\emph{Comparison with Advanced Models.} To further assess the performance of TM$^2$, we compared it against two state-of-the-art models: DMA-Nets and Seq2Event .DMA-Nets, proposed by Ji\cite{ji2021dynamic} in 2021, and Seq2Event, developed by Simpson\cite{simpson2022seq2event} in 2022, have both achieved excellent predictive accuracy in their respective fields. As shown in TABLE \ref{tab5}, TM$^2$ demonstrates superior performance in terms of MSE and MAE when compared to both DMA-Nets and Seq2Event, underscoring its development potential and room for further improvement.

\subsection{Factors Influencing the Model}
While TM$^2$ achieves strong results across most metrics, its performance in MSE, MAE, and Accuracy is slightly outperformed by DMA-Nets and Seq2Event in certain cases. Through analysis, we attribute this to the presence of dynamic change points in the dataset. The model detects and processes these change points by marking and removing data points with sudden variations and then rearranging the remaining data. Afterward, dynamic weighting is applied, which can lead to the removal of some data points that may carry valuable information—such as a sudden loss after a winning streak or unexpected player injuries. The removal of such informative points may result in slight deviations in the final results. This is likely one of the reasons why TM$^2$ did not yield larger improvements in some metrics. We hypothesize that modifying the change point detection order or fine-tuning the dynamic weighting process could further enhance the model’s performance, though we have not tested these modifications in this study.

\section{conclusions}\label{sec5}
Accurate sports prediction is essential for professional coaches, aiding in the formulation of effective training strategies and scientific competition tactics. Traditional methods, which rely on complex mathematical and statistical techniques, often face limitations due to dataset scale and struggle with long-term predictions involving variable distributions. These methods notably underperform in predicting multi-level outcomes such as point-set-game sequences.

\begin{table}[t]
\fontsize{10pt}{12pt}\selectfont
\caption{Comparison of TM$^{2}$ with state-of-the-art models}
\label{tab5}
\centering
\begin{tabular}{c|ccc}
\toprule
Model  & DMA-Nets & Seq2Event &  TM$^{2}$ \\ \midrule
MSE   & 16.8732   &  1.7476 & 0.0671 \\ 
MAE   & 10.6371   &  1.2734 & 0.0389  \\ \bottomrule
\end{tabular}
\end{table}

To address these challenges, this paper presents TM$^2$, a TCDformer-based Momentum Transfer Model for long-term sports prediction. TM$^2$ introduces a novel approach that incorporates a momentum encoding module and a prediction module based on momentum transfer. The model first encodes momentum in large-scale unstructured time series using the Local Linear Scaling Approximation (LLSA) module. Subsequently, it decomposes the reconstructed time series via momentum transfer into trend and seasonal components. The final predictions are generated by an additive combination of a Multilayer Perceptron (MLP) for trend components and a wavelet attention mechanism for seasonal components.

Comprehensive experimental results on the 2023 Wimbledon men’s tournament dataset demonstrate that TM$^2$ significantly outperforms existing sports prediction models. The proposed model achieves a reduction of 61.64\% in Mean Squared Error (MSE) and 63.64\% in Mean Absolute Error (MAE), setting a new benchmark in the field of sports event prediction.

\bibliographystyle{unsrt}
\bibliography{Match}

\end{document}